\documentclass[10pt, a4paper]{article}

\usepackage{lrec-coling2024} 
\usepackage{soul}

\title{PRODIS - a speech database and a phoneme-based language model for the study of predictability effects in Polish} 

\name{Zofia Malisz$^1$, Jan Foremski$^2$, Małgorzata Kul$^2$}  

\address{$^1$KTH Royal Institute of Technology, Stockholm, Sweden\\ $^2$Adam Mickiewicz University, Poznań, Poland \\
         malisz@kth.se, janfor@st.amu.edu.pl, kgosia@amu.edu.pl \\}

\abstract{
We present a speech database and a phoneme-level language model of Polish. The database and model are designed for the analysis of prosodic and discourse factors and their impact on acoustic parameters in interaction with predictability effects. The database is also the first large, publicly available Polish speech corpus of excellent acoustic quality that can be used for phonetic analysis and training of multi-speaker speech technology systems. The speech in the database is processed in a pipeline that achieves a 90\% degree of automation. It incorporates state-of-the-art, freely available tools enabling database expansion or adaptation to additional languages. 
 \\ \newline  }

\begin{document}

\maketitleabstract


\section{Introduction}

We present PRODIS, a speech database of contemporary Polish. The database includes a phoneme-level language model based on a light GPT architecture by \citet{nanogpt} trained on Polish Wikipedia text. The model was trained in order to provide estimates of contextual predictability, such as e.g.: surprisal, for any phoneme. While PRODIS is suitable for general purpose phonetic analysis, the language model is included to enable the first large scale study of contextual predictability effects on acoustic distinctiveness in Polish speech. 

Studies have shown that lower acoustic distinctiveness manifested in e.g. phonetic reduction (temporal and spectral) is observed in predictable contexts. And vice versa, phonetic expansion is observed when the sound in question appears in an unpredictable context~\cite{jaeger2017signal}. Contextual predictability is usually operationalised by surprisal~\cite{malisz2018surpisal}. Such encoding of information in the acoustic signal is a well-attested property in speech~\cite{aylett2006smooth2}.

There are, however, cross-linguistic differences in the magnitude and application of probabilistic effects that depend on the specific prosodic system~\cite{malisz2018surpisal, athanasopulou2017intrspch, turnbull2015pred}. PRODIS is built to study language-specific behaviour of acoustic variables under surprisal in Polish, a language with a weak acoustic expression of word stress within a very predictable stress system. 

\begin{table*}[!t]
\begin{center}
\includegraphics[scale=0.8]{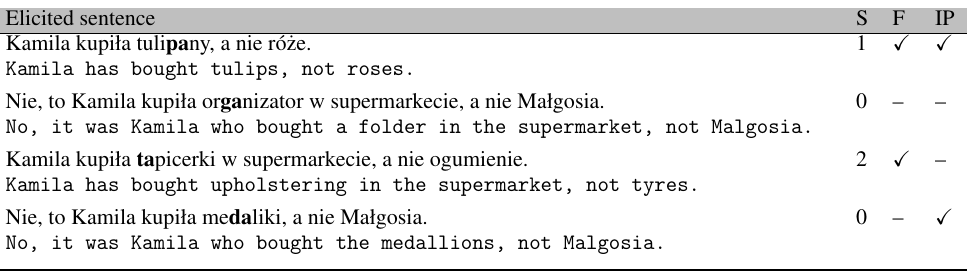} 
\caption{PRODIS prosodic task sample: lexical stress position (S: primary = 1, secondary = 2, unstressed = 0) is varied for target syllables (voiceless onsets /pa/, /ta/, /ka/; voiced onsets /ba/, /da/, /ga/) in real Polish words. Focus (F) and IP boundary (IP) conditions are elicited by embedding in meaningful sentences in a discourse completion task (context sentences not presented; target syllables were not boldfaced in the actual experiment).}
\label{tab:dct}
\end{center}
\end{table*}

Moreover, effects of predictability on acoustic distinctiveness are known to interact with prosodic structure and discourse style \cite{clopper2018assessing}. Therefore, PRODIS differentiates between discourse types: read and conversational speech. Plus, session participants engage in a speech production task in which prosodic conditions such as focus, boundary, stress are controlled.

In terms of processing, we provide fully aligned phonemic annotation for the speech data. We extract acoustic variables such as duration, $f_0$, formant values as well as spectral energy measures on the phoneme level. These are relevant for the study of prosody, discourse, predictability effects and interactions thereof.

Additionally, PRODIS fills an important infrastructural gap. Hitherto publicly available Polish speech resources have not fulfilled each of the following criteria - that PRODIS provides\footnote{Cf.   SpokesBiz \cite{spokes2023}, a speech database of ca. 500 hrs of transcribed Polish podcasts, presentations and conversations recorded via video conference software.}:

\begin{itemize}
\item it is of large size, wherein "large" concerns both the recording length and the number of speakers. At the date of publication, the database is large enough for robust statistical analysis (50 speakers, 50 minutes of speech each). The corpus has been continuously expanded since, with the aim to reach data volumes useful in the training of e.g.: data-driven multi-speaker speech synthesis. 
\item it contains a significant proportion of spontaneous, conversational speech in addition to experimentally-controlled and read speech. Hence, it includes ecologically valid, natural speech variation.
\item it is of excellent acoustic quality suitable for robust acoustic phonetic analysis and applicability in speech technology.
 It includes rich and verified annotation prepared with the purpose of acoustic phonetic incl. prosodic research.
 \item additionally, PRODIS provides a largely automated corpus creation pipeline: from recruitment and surveying of speakers, to text and speech processing and extraction of acoustic parameters. The pipeline is amenable to the construction of analogous speech corpora.
\end{itemize}

\section{Corpus architecture}
\label{sec:append-how-prod}
\subsection{Size and speaker sample characteristics}
The database contains recordings of 50 educated, native speakers of standard Polish, 25 males and 25 females who represent localities such as Warsaw, Poznań and Kraków. 
Participants were screened via a questionnaire before selection. We included adult, native speakers of Polish of good hearing and no speech impairments. We also collected standard linguistic background data on any dialects or languages spoken etc. and stored it anonymously in a JSON database. 

\subsection{Recording setup and procedure}
The participants were recorded at a speech laboratory in a fully soundproofed booth. The speakers used an AKG C 544 L head-mounted condenser microphone. Speech was recorded using a FocusRite 2i2 Scarlett audio interface and Audacity (v.3.2) running on a MacBook Pro 14 (MacOS Sonoma) at a 44100Hz sampling frequency. An auditory screening of all the recordings was performed. Recordings failing to meet the criterion of excellent acoustic quality were excluded.

The participants signed a GDPR-compliant consent in which they were informed about anonymity, ethical provisions and the right to withdraw data. We provided an option to agree to data use for training Polish non-commercial TTS. We explained the potential use of the TTS voice. 

The experimenter (not miked) and the participants were separated by a glass window and maintained eye contact during the recording. Speakers were remunerated for their time (40-90 min.) by means of a gift card and were offered a debrief about study intent after the recording.

\subsection{Database design}

The database involves two discourse-differentiated conditions: a conversation and a reading task as well as a separate prosodic production task. The database task design is presented in Fig. \ref{fig:pipeline} in orange boxes. The participants navigated the tasks and task items themselves using a website \url{https://prodis-opus19.github.io/experiment/}. The task order as well as the item order within a task was automatically randomized per participant. 

\subsubsection{Conversation and reading}
The conversation condition entailed a spontaneous interaction between the experimenter and the participant about their favourite experience such as a party, a festival or a family event. The interaction took between 10 and 20 min. per speaker.

The reading task consisted of reading four texts drawn from Wikipedia. There were three sets of four texts counterbalanced per participant. We eliminated mathematical formulas, problematic proper nouns, and other elements that might potentially hinder the accuracy of speech processing, from the texts. The total word count per text was ca. 350 words, with unique word count on average ca. 250 words. The texts represented four topical domains: politics, 
science, history 
and culture. 

\subsubsection{Prosodic task}

The prosodic task followed a discourse completion task  procedure used in~\cite{malisz2018stress}: participants were asked to listen to a pre-recorded context sentence in Polish and to read the stimulus sentence in response. Table \ref{tab:dct} presents a sample of the study design and materials. Participants produced syllables of the form C+/a/ in real Polish words (lemmata N=42, no compounds) in several stress, focus and boundary conditions. Two sets of words were used: four- and five-syllables long, eliciting secondary stress that in Polish falls on the initial in words longer than three syllables. Each target syllable occurred once in the penultimate position for primary stress, in the initial position for secondary and in other, non-final position for unstressed. Each target word was embedded in carrier sentences exemplified in Table \ref{tab:dct}, varying IP boundary (medial vs. final) and focus. One repetition per factor level combination was recorded per speaker resulting in over 6000 produced samples.

\begin{figure*}[!t]
\begin{center}
\includegraphics[scale=0.23]{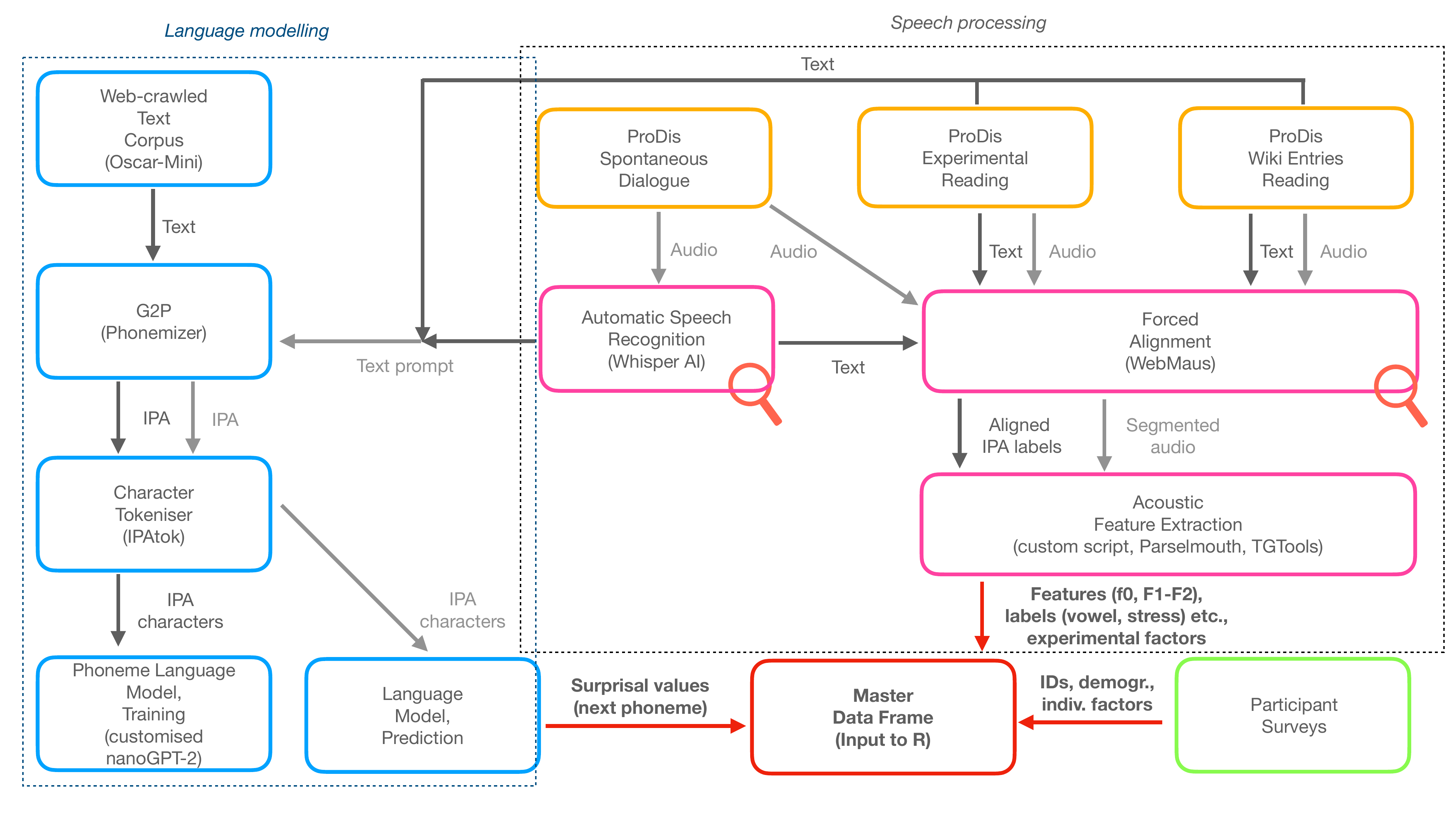} 
\caption{PRODIS: database design (orange), speech processing (pink) and language modeling (blue) pipelines. The magnifying glass symbolises manual verification processes.}
\label{fig:pipeline}
\end{center}
\end{figure*}

\section{Speech processing}
\label{sec:speechprocessing}

We processed the corpus using out-of-the-box tools achieving a 90\% level of automation. The speech processing pipeline is presented in Figure \ref{fig:pipeline} in pink boxes.

\subsection{ASR and forced alignment}
The speech recorded in the spontaneous conversation task required transcription. We used a data-driven neural model for Polish implemented as part of the WhisperAI ASR platform~\cite{whisper}. Our pipeline accepts a single input directory before proceeding with transferring mono audio and the corresponding text transcriptions generated by the Whisper model, into a single output directory. Naming scheme and directory structure were preserved and errors logged. The time required to transcribe a single recording on an RTX 3060 Ti was marginally longer than the actual duration of the recording itself, for example, 18 minutes of processing time for a 15-minute recording.

The resulting transcriptions were subsequently examined manually. The team, on average, spent one hour evaluating a conversation transcription. While the Polish WhisperAI model achieves remarkable accuracy, it tends to be prescriptive in correction of speech errors such as hesitations, repetitions, false starts or vocalizations. We had to reinsert such phenomena into the transcription manually to facilitate phonemic forced alignment.  

The transcriptions for all tasks and original audio files were then used for forced alignment using WebMAUS ~\cite{kisler2017multilingual}. Text normalisations (numbers, abbreviations) on all input texts were introduced manually. To align, we used a WebMAUS Pipeline method with Chunker and WebMAUSBasic in the processing chain. 
The software received .wav and .txt file (standard Polish orthography) pairs as input and generated a time-aligned Praat TextGrid ~\cite{Praat} with word and phonemic segmentation and IPA labels. The alignment was manually verified by the team. 

\subsection{Acoustic feature extraction}
We automatically extracted several acoustic features relevant for phonetic and prosodic analysis from the recorded speech. We used TextGridTools ~\cite{buschmeier2013textgridtools} and Parselmouth~\cite{parselmouth} Python libraries to process Praat TextGrid files and call Praat signal processing functions, respectively. The extraction pipeline is customisable and expandable. So far the following feature implementations are available:

\begin{itemize}
\item \textbf{Fundamental frequency}: $f_0$ values were extracted from vowel phonemes with a time step of 0.01~s and using a pitch floor and ceiling of 75~Hz and 300~Hz respectively. 
\item \textbf{Formant frequencies}: we used the Burg method in Praat with a time step of 0.01s, a 0.025s long analysis window and the maximum number of formants set to 5. Pre-emphasis was applied over 50Hz. The formant ceilings were initialised to 5000Hz for male speakers and 5500Hz for female ones and optimised following~\citet{escudero2009cross}. 
\item \textbf{Fundamental and formant frequency contours}: 
we length-normalised the contours per each vocalic segment over nine evenly distributed observation points, excluding formant transition points. Formant contours were normalised per speaker following~\citet{lobanov1971classification}; $f_0$ contours were normalised to zero mean and unit variance relative to each speaker.
\item \textbf{Spectral Tilt}: extracted as the predictor coefficient of the first-order all-pole polynomial~\cite{raitio2022hierarchical}.
\item \textbf{Spectral Centroid}: extracted as the average of frequency bin values weighted by their corresponding frequency value.
\end{itemize}

Further features currently in implementation involve: Voice Onset Time for stops, spectral center of gravity for fricative consonant portions, as well as further prosodic measures incl. discrete and continuous labelling of prosodic prominence levels.

\section{Phoneme-level language modelling}
\label{sec:languagemodel}
We converted a web-crawled Polish text corpus into IPA and trained a phoneme-level GPT2-equivalent language model on this data~\cite{nanogpt}. This way, more accurate predictions, closer to speech production processes can be obtained, compared to a regular word-level model \cite{wilcox2020predictive}. 
The language modelling pipeline is presented in Figure \ref{fig:pipeline} in blue boxes.

We trained the model on OSCAR-mini~\cite{ortiz-suarez-etal-2020}, a downscaled (50 MB in Latin, ~80 MB after IPA phonemisation) variant of OSCAR~\cite{oscar}, both un-annotated multilingual datasets based on the Common Crawl. 

Due to the requirement to use phoneme-level transcriptions instead of the Latin alphabet, it was not feasible to utilize a pre-trained character-level model. We used the Phonemizer~\cite{Bernard2021}, a Python library, to machine-convert characters to Polish phoneme labels. Phonemizer also removes non-standard characters e.g.: emojis during the conversion process. We created a simple wrapper class around Phonemizer with predefined settings for Polish. Phonemizer itself offers multiple backends, we chose E-speak, as it removes foreign language strings automatically.

Our tokeniser is based on ipatok~\cite{ipatok}. Ipatok is a Python library that splits IPA strings at individual IPA phonemes. It does not include spaces, so we added space separators. We also added multiprocessing, improving speed 5-fold.

As a base for language modeling we adopted Nano-GPT~\cite{nanogpt}, a simple and fast model for training and fine-tuning medium-sized GPTs. To train on OSCAR-mini, we used the default NanoGPT configuration: 12 layers and heads, a size embedding of 768 with zero dropout and the AdamW optimizer. We modified the following parameters: batch size of 64, with a context of up to 256 previous tokens in a block. The learning rate was 1e-4 (min. of 1e-5). 
We trained on a single Nvidia RTX 3060 Ti 8GB GDDR6X and yielded the loss value of 0.47 at the max. iteration of 60 000 with 85M number of parameters. We evaluated the model on a validation set every 250 batches with early stopping after 5 evaluations. We split the data into 90\% train-10\% development sets, using IPA tokens as delimiters. 



After training, we sampled the language model to obtain probabilities for each phoneme read or spoken in the speech database. We fed prompts with a rolling window of 10-token length. We returned the intermediary tensor containing the whole predicted distribution of probabilities for the next token (= 66 IPA and special characters), instead of returning the single most probable token. 

We use the probabilities to calculate surprisal for the next phoneme that actually occurs in the speech database. Surprisal quantifies the
amount of information (in terms of \textit{bits}) as the inverse of the unit’s
\textit{log} probability given a local context:

\begin{equation}\label{eq:id}
Surprisal(unit_i) = -log_2 P(unit_i|Context)
\end{equation}

Depending on the analytical goal of future studies using PRODIS, other information-theoretic measures known to affect speech can be derived from the predicted probabilities, for example, informativity~\cite{cohen2017len} or entropy~\cite{shaw2019effects}.

\section{Conclusion}

We presented PRODIS a Polish speech database for the study of discourse, prosodic and predictability effects on a host of acoustic parameters. PRODIS is a valuable addition to the Polish linguistic resource landscape as a fully-aligned phonetic speech corpus that also includes a phoneme-level language model useful for investigating information-theoretic effects on speech. We intend to use the design, pipeline and tools to record further databases of under-researched Slavic languages in the near future.


\section{Acknowledgements}

We gratefully acknowledge the financial support of the National Science Center (OPUS grant, project no. 2020/37/B/HS2/04161). We thank Cyprian Nosek for valuable input regarding the language model.

\section{Bibliographical References}\label{sec:reference}

\bibliographystyle{lrec-coling2024-natbib}
\bibliography{lrec-coling2024-example}


\end{document}